\documentclass[runningheads,a4paper]{llncs}
\usepackage{amsmath}
\usepackage{amssymb}
\usepackage{amsfonts}
\usepackage{amssymb}
\usepackage{graphicx}
\usepackage{subfigure}
\usepackage{breakcites}
\usepackage{nicefrac}
\usepackage{color}
\usepackage{bm}
\usepackage{cite}
\usepackage{url}
\usepackage{stackrel}
\usepackage{float}
\usepackage[inline]{enumitem}

\definecolor{darkgreen}{rgb}{0,0.6,0.2}

\newcommand{\keywords}[1]{\par\addvspace\baselineskip
\noindent\keywordname\enspace\ignorespaces#1}

\begin{document}

\mainmatter  

\title{A clustering approach to heterogeneous change detection}

\titlerunning{A clustering approach to heterogeneous change detection}

%
\author{Luigi Tommaso Luppino%
\thanks{luigi.t.luppino@uit.no},
Stian Normann Anfinsen,
Gabriele Moser,
Robert Jenssen,
Filippo Maria Bianchi,
Sebastiano Serpico,
Gregoire Mercier
}
\authorrunning{Luppino et al.}

\institute{Machine Learning Group, Dept. of Physics and Technology, University of Troms\o{}, Norway}

%
%

\toctitle{Title of the paper}
\tocauthor{Luppino et al.}
\maketitle

\begin{abstract}
Change detection in heterogeneous multitemporal satellite images is a challenging and still not much studied topic in remote sensing and earth observation. 
This paper focuses on comparison of image pairs covering the same geographical area and acquired by two different sensors, one optical radiometer and one synthetic aperture radar, at two different times. 
We propose a clustering-based technique to detect changes, identified as clusters that split or merge in the different images.
To evaluate potentials and limitations of our method, we perform experiments on real data.
Preliminary results confirm the relationship between splits and merges of clusters and the occurrence of changes.
However, it becomes evident that it is necessary to incorporate prior, ancillary, or application-specific information to improve the interpretation of clustering results and to identify unambiguously the areas of change.
\end{abstract}

\section{Introduction}
\label{sec:intro}
Change detection systems provide crucial information for damage assessment after natural disasters such as floodings, earthquakes, landslides, or to detect long-term trends in land usage, urban development, glacier dynamics, deforestation, and desertification \cite{serpico2012information,dell2012remote,ningning2012landslide,du2012spatial,wu2011glacier,latif2015deforestation,jianjun2004theory}. 
In the last years, thanks to the development of heterogeneous or multimodal change detection methods, it was possible to relax the assumption of homogeneous and co-calibrated measurements.
However, despite its undeniable potential, there is still a limited amount of research on heterogeneous change detection in the fields of computer vision, pattern recognition and machine learning.
In \cite{mercier2007conditional}, copula theory is exploited to build local models of dependence between unchanged areas in heterogeneous images and to link their statistical distributions.
In \cite{storvik2009combination}, joint distributions of heterogeneous images are obtained by transforming their marginal densities in meta-Gaussian distributions, which provide simple and efficient models of multitemporal correlations. 
In \cite{liu2012dynamic,liu2014change}, a method based on evidence theory is proposed, which fuses clustering maps of the individual heterogeneous images and then detects "change" and "no-change" classes, from the transition probabilities between clusters. 
In \cite{prendes2015new}, the physical properties of the considered sensors and, especially, the associated measurement noise models and local joint distributions are exploited to define a "no-change" manifold.

The capability of processing data from heterogeneous sources in the same application opens for usage of a much larger amount of information. 
With respect to time series, the temporal resolution can be increased and the overall time window can be extended.
Nonetheless, new issues arise. 
Different sensors are sensitive to distinct physical conditions and comparing their measurements may produce false detections, due to inconsistencies in sensor behaviour rather than actual changes in the monitored entities. 
As the complexity of the fused data set increases, there could be a requirement for more flexible and complicated statistical models, which are harder to fit on data, they may be characterized by larger uncertainty in the parameter estimation and a higher computational cost. 
Finally, detecting and characterizing changes in heterogeneous images is not as trivial as in the homogeneous case, where a change corresponds simply to a difference in the signal values.

In this work, we propose a novel cluster-based approach for change detection in heterogeneous data.
We design an unsupervised method to be as general as possible, i.e.\ application-independent.
The proposed method processes pairs of images, acquired at different times from different sensors. 
In particular, one image comes from an optical sensor, whereas the second is a synthetic aperture radar (SAR) image. 
The images must be co-registered by a pre-processing step, to avoid that spatial misalignment of the images is misclassified as a change. Moreover, a third type of images is considered, whose elements are obtained by stacking optical and SAR images.
A clustering method is executed independently on each of the three data sets.
Then, the clusters identified in the first two data sets are matched against the ones from the third data set, in order to determine if the clusters from the first image split or merge in the second image.
We associate changes to the occurrence of such modifications.

In this preliminary study, the problem has been defined, a possible solution has been suggested and experiments have been performed to assess the capability of the proposed methodology. 
Making the whole process automatic is the following step, which will be treated in a further extension of this work.

\section{Background}
\label{sec:background}

This work leverages on the information delivered by distance-based clustering analysis on image data. 
To select the proper distance measures, we first need to identify the correct statistical models to represent the data. 
Since we process optical and SAR images, we consider only models commonly used when dealing with these two specific data. 
 
A simple probability distribution that describes well the optical images is the Gaussian distribution \cite{bovolo2015time,goudail2004contrast}.
Specifically, a sensor with $n$ channels yields feature vectors $\boldsymbol{x}_{opt} \in \mathbb{R}^n$, which are modelled by a multivariate Gaussian probability density function (pdf)
\begin{equation*}
    f(\boldsymbol{x}_{opt}|\boldsymbol{\mu}_{i},\boldsymbol{\Sigma}_{i})=\frac{1}{(2\pi)^{\nicefrac{n}{2}}|\boldsymbol{\Sigma}_{i}|^{\nicefrac{1}{2}}}\exp\left(-\frac{1}{2}(\boldsymbol{x}_{opt}-\boldsymbol{\mu}_{i})^{t}\boldsymbol{\Sigma}^{-1}(\boldsymbol{x}_{opt}-\boldsymbol{\mu}_{i})\right)\,,
\end{equation*}
which compactly reads as $\boldsymbol{x}_{opt}|\omega_{i}\sim N(\boldsymbol{\mu}_{i},\boldsymbol{\Sigma}_{i})$. Here $\boldsymbol{\mu}_{i}$ and $\boldsymbol{\Sigma}_{i}$ are the mean vector and the covariance matrix associated to cluster $\omega_{i}$, respectively. 

Concerning SAR images in single polarisation, using the gamma distribution is a simplistic, yet effective option \cite{oliver2004understanding}:
\begin{equation*}
    f(x_{SAR}|\theta_{i},L)=\frac{1}{\theta_{i}\Gamma(L)}\left(\frac{x_{SAR}}{\theta_{i}}\right)^{L-1}\exp\left(-\frac{x_{SAR}}{\theta_{i}}\right)\,.
\end{equation*}
This is denoted by $x_{SAR}|\omega_{i}\sim \Gamma(\theta_{i},L)$. 
$\varGamma(L)$ is the gamma function, while $L$ and $\theta_{i}$ are the shape and the scale parameter, respectively. 
Since $L$ (the number of looks) is the same for all the clusters, these can be fully characterised by their mean $\mu_{i}=L\theta_{i}$.

The log-normal distribution is an alternative to the gamma pdf. 
It fits data reasonably well under most circumstances and, contrarily to the gamma pdf, it allows to model heavy-tailed SAR intensity data \cite{oliver2004understanding}. 
A positive-valued random variable $X|\omega_{i}=e^{Y}$ follows a log-normal distribution if $Y|\omega_{i}=\log(X)\sim N(\mu_{i},\sigma_{i})$. 
The pdf reads
\begin{equation*}
    f(X|\mu_{i},\sigma_{i})=\frac{1}{X\sqrt{2\pi\sigma_{i}^{2}}}\exp\left(-\frac{(\log(X)-\mu_{i})^{2}}{2\sigma_{i}^{2}}\right)\,,
\end{equation*}
denoted by $X|\omega_{i}\sim logN(\mu_{i},\sigma_{i})$. 
The first two moments of random variables $X$ and $Y$ are related according to
\begin{align*}
    \mu_{X|\omega_{i}}	&=	\exp\left(\mu_{i}+\frac{\sigma_{i}^{2}}{2}\right),\;\;
    \sigma_{X|\omega_{i}}^{2} =\mu_{i}^{2}\left(e^{\sigma_{i}^{2}}-1\right).
\end{align*}

To conclude, if the statistical behaviour of a SAR image can be described by log-normal distributions, then a logarithmically transformed image can be modelled by a Gaussian distribution. 
This property will be useful to process the stacked data $\boldsymbol{x}_{st}$, which combines all features of the optical and the SAR image into one stacked feature vector, associated to each pixel. 

As distance measures, we use Mahalanobis distance \cite{basseville1989distance} for multivariate Gaussian distributed data and Hellinger distance \cite{frery2014analytic} for gamma distributed data. 
A notorious drawback in cluster methods is the dependence of their results to initial conditions, such as initialization of cluster centers and ordering of the data. 
Additionally, the desired number of clusters or the scale parameter (used in methods such as hierarchical or density-based clustering) is often unknown.
Ensemble clustering methods tackle these issues, by providing more stable results at the cost of higher computational complexity \cite{strehl2002cluster,fred2005combining,ghosh2011cluster}.
Ensemble methods can identify clusters of nontrivial shape and with different densities, handle noise and outliers, and they provide an estimate to the optimal number of clusters.
In our case, such a number is unknown and, therefore, we perform cluster analysis with an ensemble approach based on Fuzzy C-Means (FCM) \cite{ghosh2009unsupervised,singh2013unsupervised,keller1985fuzzy}.
The ensemble procedure consists in repeating several times the FCM initialized with $k$ different number of clusters, which each time is drawn from a uniform discrete distribution.
FCM is implemented with the distance measures mentioned above. 
The FCM algorithm represents an iterative approach, where at each iteration a partition matrix $U$ is returned as output. The membership values $\mu_{ij}$ contained in $U$ are exploited to evaluate the covariance matrix of each cluster as:
\begin{equation*}
    \boldsymbol{\Sigma}_{i}=\frac{\stackrel[j=1]{N}{\sum}\mu_{ij}(\boldsymbol{x}_{j}-\boldsymbol{c}_{i})(\boldsymbol{x}_{j}-\boldsymbol{c}_{i})^{t}}{\stackrel[j=1]{N}{\sum}\mu_{ij}}, \quad i=1\,,\,\ldots\,,\,k.
\end{equation*}

When multivariate Gaussian distributed data are involved, the Mahalanobis distance computed in the following iteration employs these updated covariance matrices. 
In a possible future development, we plan to examine the partition matrix to identify the most reliable clustering results, in order to improve the post-clustering analysis.


\section{Recognition of cluster splits and merges}
\label{sec:methods}

\begin{figure}[ht!]
\centering
    \includegraphics[width=0.6\textwidth]{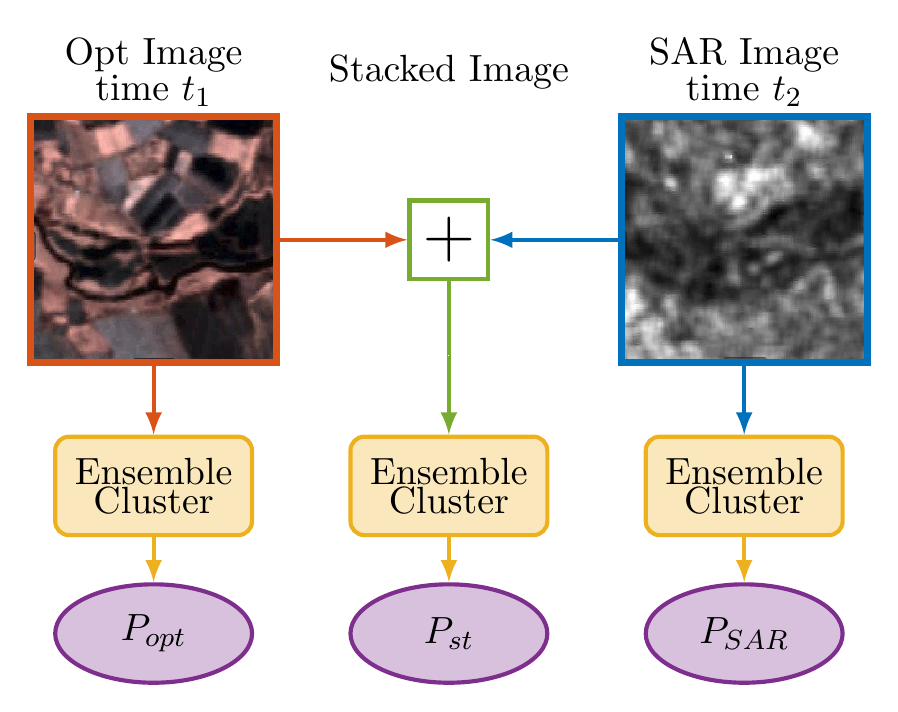}
\caption{First step of the proposed methodology: obtainment of the stacked image and of the three partitionings.}
\label{fig:3clustering}
\end{figure}

Given two heterogeneous images of the same geographical area captured respectively at times $t_1$ and $t_2$, we want to detect if a change occurred during the time lapse.
Each image is clustered by using the distance measures that captures its statistical properties. 
The clustering ensemble procedure on each image provides the partitions
\[P_{opt}=c_{opt}^{(1)} \cup \quad \dots \quad \cup c_{opt}^{(N_{opt})}\] \[P_{SAR}=c_{SAR}^{(1)} \cup \quad \dots \quad \cup c_{SAR}^{(N_{SAR})}
\]
where $N_{opt}=\left|P_{opt}\right|$ and $N_{SAR}=\left|P_{SAR}\right|$ are the number of clusters in each partition.
Then, if the SAR data $x_{SAR}$ are assumed to follow a log-normal distribution, the logarithm of their intensities can be modelled by Gaussian pdfs.
Since also the optical data $\boldsymbol{x}_{opt}$ are modelled by Gaussian pdfs, the stacked vector $\boldsymbol{x}_{st}=\left[\boldsymbol{x}_{opt} \,,\log\left(x_{SAR}\right)\right]$ could be thought of a realization of a multivariate Gaussian random variable. 
Accordingly, we compute a third partition $P_{st}$ on the stacked data, as shown in Fig.\ \ref{fig:3clustering}.
To determine the number of clusters to be considered by the ensemble procedure on stacked images, useful information can be extracted from the clustering results obtained on optical and SAR images. 
In fact, the allowed number of clusters provided to every instance in the ensemble procedure for the stacked image are drawn from the uniform discrete pdf $U \left[ \max \left(N_{opt},N_{SAR}\right), N_{opt}\cdot N_{SAR}\right]$.

\begin{figure}[ht!]
\centering

    \includegraphics[width=\textwidth]{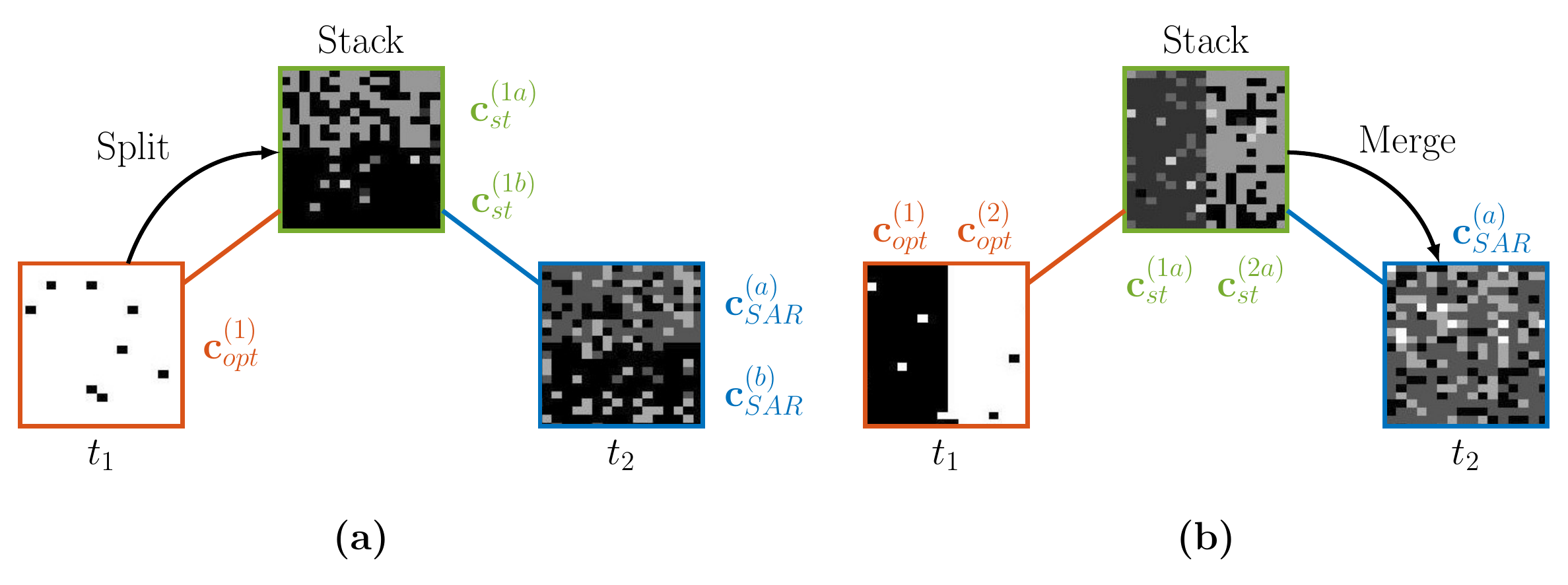}
		
\caption{Proposed methodology: it is possible to recognize changes as splits (a) and merges (b) by the comparison with the partitioning of the stacked image.}
\label{fig:splitmerge}
\end{figure}

Once the three partitions are obtained, we check whether a cluster from the image at time $t_{1}$ splits into two or more clusters in the stacked image, or whether two or more clusters from the stacked image may merge into one cluster of the image at time $t_{2}$. Instead of comparing directly the clusters from time $t_{1}$ and time $t_{2}$, with our method we leverage the information contained in the covariance matrix of the stacked image, which captures the cross-correlation between the original images. 
Moreover it may provide a regularization that filters out the effect of the speckle noise on the clustering results. 
The proposed methodology is depicted in Fig.\ref{fig:splitmerge}. 
In Fig. \ref{fig:splitmerge}(a), a region in the optical image at time $t_1$ is fully contained in a cluster $\mathbf{c}_{opt}^{(1)}$. The same region, is divided in two clusters, $\mathbf{c}_{st}^{(1a)}$ and $\mathbf{c}_{st}^{(1b)}$, in the stacked image. 
This means that in the SAR image at $t_2$ the region is split in two clusters as well, $\mathbf{c}_{SAR}^{(a)}$ and $\mathbf{c}_{SAR}^{(b)}$. This denotes that a change occurred in the time lapse $t_2-t_1$.
In Fig. \ref{fig:splitmerge}(a) instead, we can see the region that in the stacked image corresponds to two clusters $\mathbf{c}_{st}^{(1a)}$ and $\mathbf{c}_{st}^{(2a)}$, merges into a single cluster $\mathbf{c}_{SAR}^{(a)}$ in the SAR image at time $t_2$. This indicates another type of change from $t_1$, where the region is characterized by two clusters $\mathbf{c}_{opt}^{(1)}$ and $\mathbf{c}_{opt}^{(2)}$ in the optical image.

\section{Experiments and results}
\label{sec:experiments}

In this section the proposed approach is applied, showing the potential and limitations of ensemble clustering and of an analysis of splits and merges.


\begin{figure}[ht!]
\centering

    \subfigure[Sept. the 5th, 1999]{
    \includegraphics[width=0.3\textwidth, height= 6cm ]{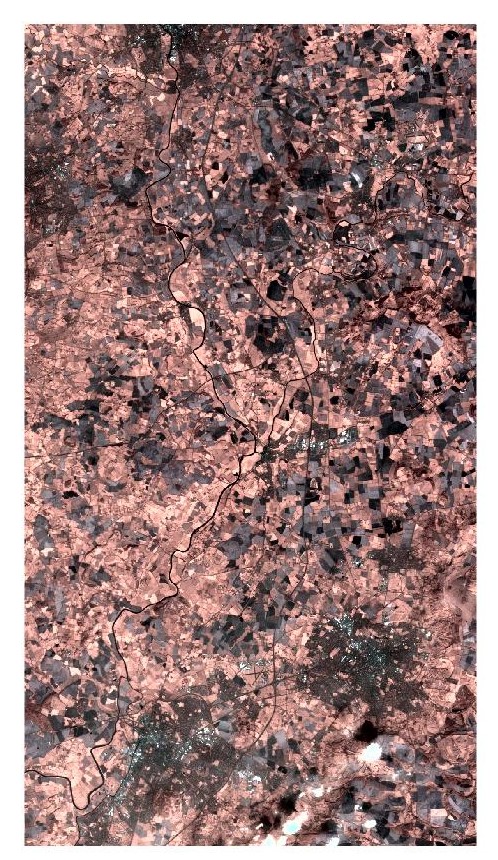}
    \label{fig:opt1}}\hspace{0em}%
    ~
	\subfigure[Oct. the 21st, 2000]{
    \includegraphics[width=0.3\textwidth, height= 6cm]{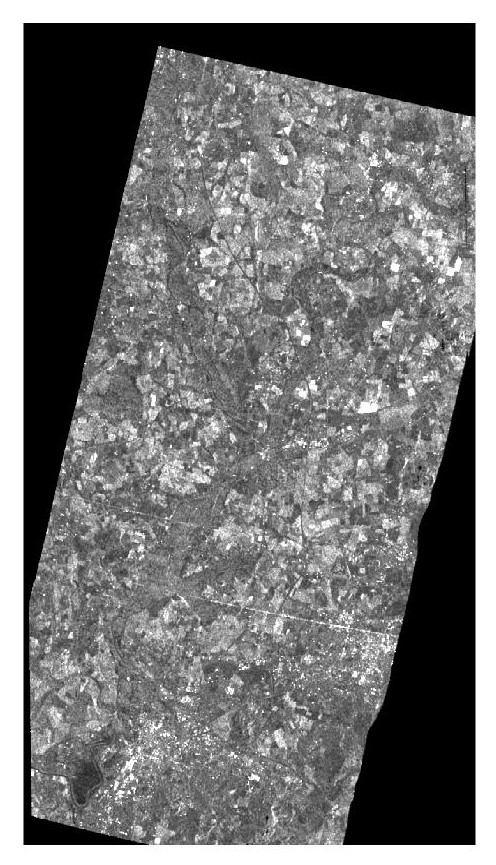}
    \label{fig:SAR2}}\hspace{0em}%
    ~
	\subfigure[Ground truth]{
    \includegraphics[width=0.3\textwidth, height= 6cm]{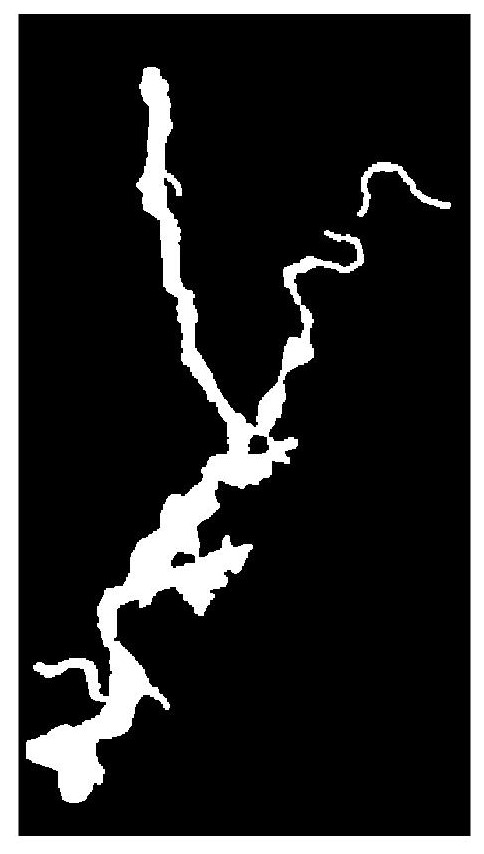}
    \label{fig:GT}}
		
\caption{Gloucester before -- optical image (a) -- and after a flooding the event -- SAR image (b). In (c), the ground truth of the change.}
\label{fig:dataset}
\end{figure}

The images in Fig.\ \ref{fig:dataset} represent the countryside at the periphery of Gloucester, Gloucestershire, United Kingdom, before and after a flood. 
Since the speckle noise affecting the latter was too strong, a 7-by-7 enhanced Lee filter \cite{lopes1990adaptive} was applied to attenuate the noise, while preserving the details contained in heterogeneous areas. 
The analysis is carried out on the presented images by dividing them into smaller and non-overlapping windows of $50 \times 50$ pixels, and then by looking for changes inside them separately. 
In this way, it can be reasonably thought that pixels can be grouped into a limited number of clusters, making the clustering process easier and more accurate regardless of the spatial nonstationarity of the image data. 
Processing smaller windows also reduces the computational cost, which scales quadratically with the windows size and the number of clusters. 
The FCM algorithm has been iterated 20 times, drawing a different number of clusters each time from the uniform probability mass function $U[4,7]$.

\subsection{First experiment}

\begin{figure}[h!]
\centering

    \subfigure[Time $t_1$]{
    \includegraphics[width=0.3\textwidth, height = 2.6cm]{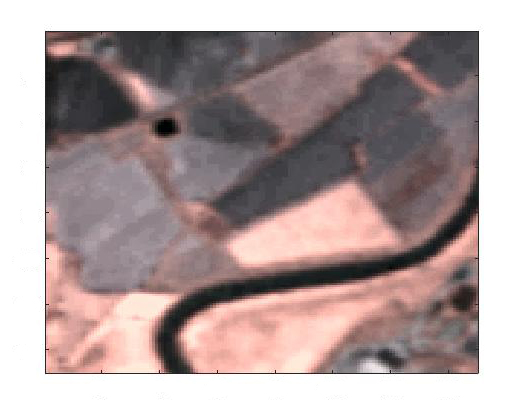}
    \label{fig:o_w1}}\hspace{0em}%
    ~
	\subfigure[Time $t_2$]{
    \includegraphics[width=0.3\textwidth, height = 2.6cm]{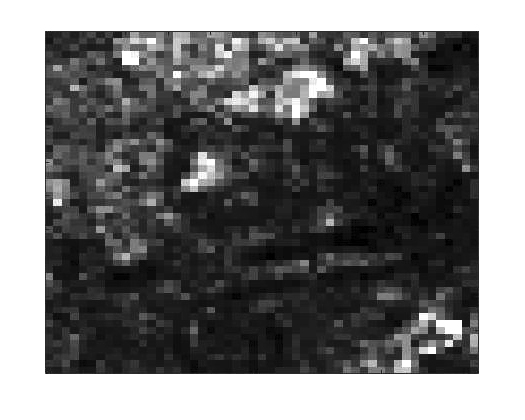}
    \label{fig:s_w1}}\hspace{0em}%
    ~
	\subfigure[Ground truth mask]{
    \includegraphics[width=0.3\textwidth, height = 2.6cm]{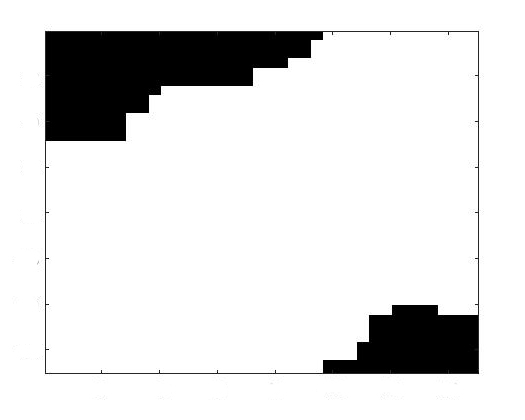}
    \label{fig:mask_w1}}\hspace{0em}%
    ~
	\subfigure[Clustering $t_1$]{
    \includegraphics[width=0.3\textwidth, height = 2.6cm]{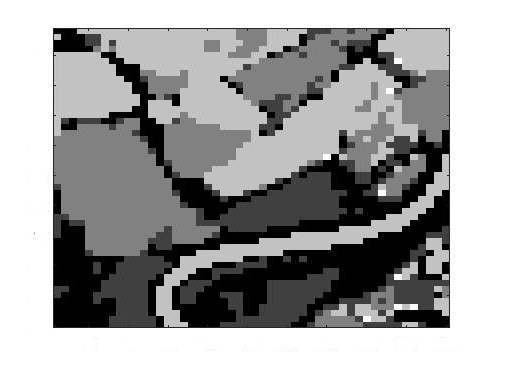}
    \label{fig:clo_w1}}\hspace{0em}%
    ~
	\subfigure[Clustering $t_2$]{
    \includegraphics[width=0.3\textwidth, height = 2.6cm]{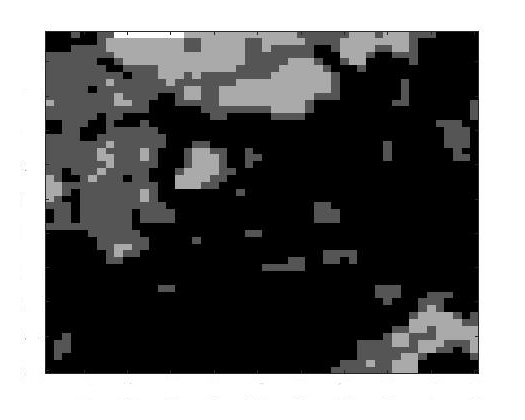}
    \label{fig:cls_w1}}\hspace{0em}%
    ~
    \subfigure[Clustering ${t_1+t_2}$]{
    \includegraphics[width=0.3\textwidth, height = 2.6cm]{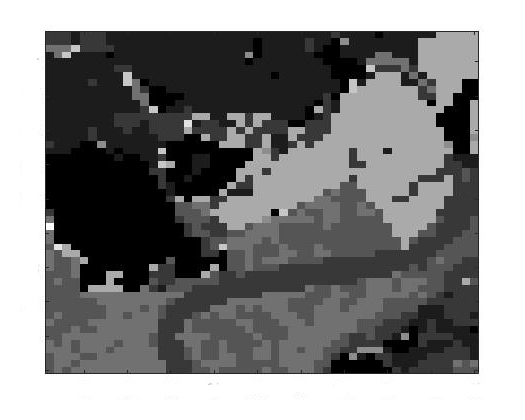}
    \label{fig:clSt_w1}}
		
\caption{First experiment: (a) window of the optical image, (b) window of the SAR image, (c) window of the ground truth mask, (d) clustering result on image a, (e) clustering result on image b, (f) clustering result on the stacked image.}
\label{fig:w1}
\end{figure}

The region selected for the first experiment is shown in Fig.\ \ref{fig:w1}. 
It contains some agricultural fields and a river in the lower part of it. 
As seen in Fig.\ \ref{fig:clo_w1}, clusters relative to different parts of the image are very well separated.

The true number of clusters is unknown and it must be properly evaluated for a correct separation of the objects. In fact, an excessive (or insufficient) number of clusters will lead to oversegmentation (or undersegmentation) of the image. 
Concerning the SAR acquisition, from  Fig.\ \ref{fig:s_w1} we observe that the flooded area covers the majority of the window.
Such area is correctly identified by the large black cluster in Fig.\ \ref{fig:cls_w1}.
Comparing the three clustering results, two clear examples of clusters merging and clusters splitting are spotted. 
The big cluster representing some fields, that from the upper part of the optical image goes down to the right, has split into two different clusters in the SAR image (the light grey one and the black one), and this is highlighted by the presence of the light grey and the dark grey clusters in Fig.\ \ref{fig:clSt_w1}. 
Then, the dominant cluster of Fig.\ \ref{fig:cls_w1} is the result of the merging of some clusters of the optical image, i.e.\ the white cluster (the river), the dark grey cluster (some fields close to the river), a good percentage of the black cluster (the boundaries around the river and the ﬁelds) and one part of the above mentioned big cluster which has split. All these clusters are visible in the result obtained with the stacked data, and they are respectively: the dark grey cluster (the river), the grey cluster close to it (the fields close to the river and the boundaries) and the light grey cluster (the part of the splitting).

\subsection{Second experiment}

\begin{figure}[h!]
\centering

    \subfigure[Time $t_1$]{
    \includegraphics[width=0.3\textwidth, height = 2.6cm]{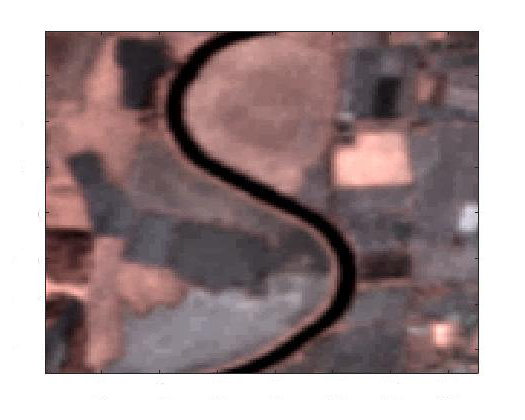}
    \label{fig:o_w2}}\hspace{0em}%
    ~
	\subfigure[Time $t_2$]{
    \includegraphics[width=0.3\textwidth, height = 2.6cm]{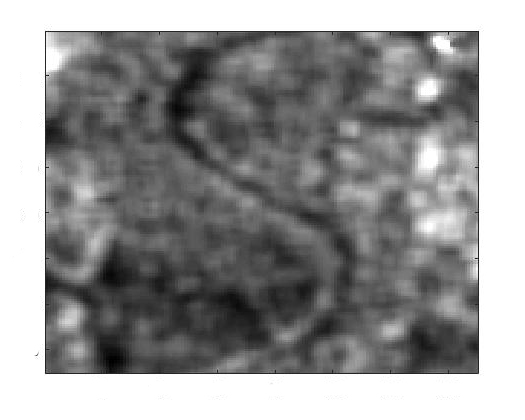}
    \label{fig:s_w2}}\hspace{0em}%
    ~
	\subfigure[Ground truth mask]{
    \includegraphics[width=0.3\textwidth, height = 2.6cm]{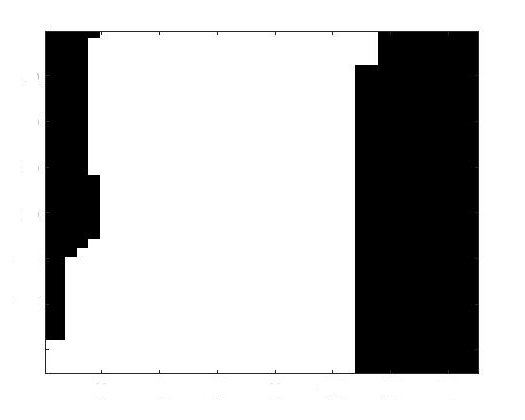}
    \label{fig:mask_w2}}\hspace{0em}%
    ~
	\subfigure[Clustering $t_1$]{
    \includegraphics[width=0.3\textwidth, height = 2.6cm]{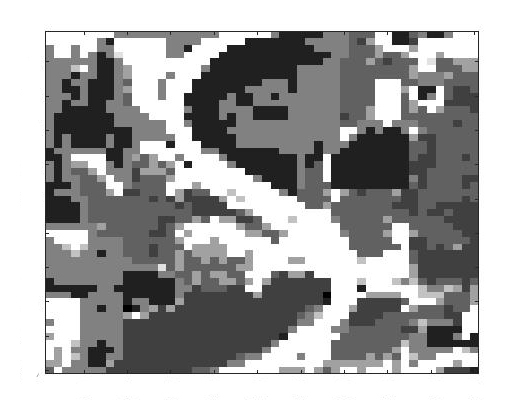}
    \label{fig:clo_w2}}\hspace{0em}%
    ~
	\subfigure[Clustering $t_2$]{
    \includegraphics[width=0.3\textwidth, height = 2.6cm]{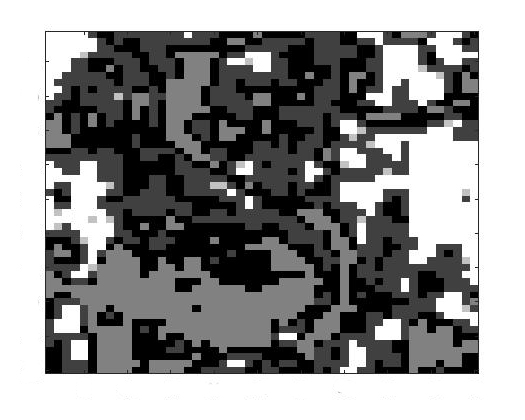}
    \label{fig:cls_w2}}\hspace{0em}%
    ~
	\subfigure[Clustering $t_1+t_2$]{
    \includegraphics[width=0.3\textwidth, height = 2.6cm]{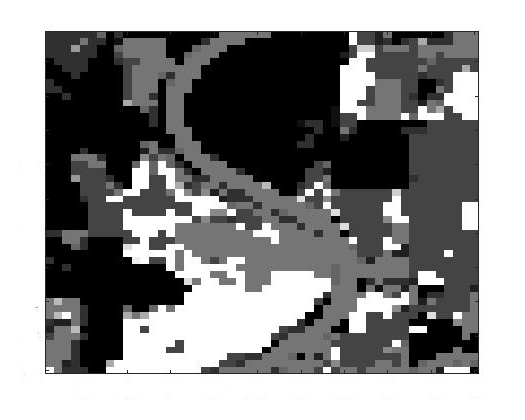}
    \label{fig:clSt_w2}}
		
\caption{Second experiment: (a) window of the optical image, (b) window of the SAR image, (c) window of the ground truth mask, (d) clustering result on image a, (e) clustering result on image b, (f) clustering result on the stacked image.}
\label{fig:w2}
\end{figure}

The region selected for the second experiment is displayed in Fig.\ \ref{fig:w2}. 
In this case, the different areas are not well separated (Fig.\ \ref{fig:clo_w2}), especially in the center and in the lower right corner of the window, mainly because these parts of the image present miscellaneous ground covers. 
For example, some of the central pixels in Fig.\ \ref{fig:o_w2} look darker, so the clustering algorithm erroneously cluster them together with the ones belonging to the river, as it happened in the ﬁrst experiment. Instead, the bare soil field presents some brighter pixels close to the river and some darker pixels far from it, and these two groups are divided. 
Moving on to the image in Fig.\ \ref{fig:s_w2}, it can be seen how it looks still noisy and muddled, even after being filtered. 
Consequently, the clustering in Fig.\ \ref{fig:cls_w2} does not yields the same quality of the first experiment. 
Making a comparison with Fig.\ \ref{fig:mask_w2}, more accurate delineation of changed areas would have been emphasized if the grey and black classes were grouped together and, most importantly, some of the agricultural fields in the lower right corner were grouped diﬀerently. 
But this is not a fault of the ensemble clustering, as these last areas are very similar to the flooded portion of the region, due to the characteristics of the specific kind of field and its SAR signature. 
Recognizing the flooded area in Fig.\ \ref{fig:s_w2} by visual inspection and without prior knowledge, is also very difficult. 
It is worth noting that the available ground truth itself is only partially accurate, because for example the sharp edge on the right side is unlikely, but it still gives an idea about the location of the affected areas.

As expected according to the aforementioned characteristics of the SAR scene, Fig.\ \ref{fig:clSt_w2} shows how the output of the ensemble clustering is irregular. The quality of the partitioning is heavily influenced by the speckle noise, which is a fundamental issue in the field of SAR data analysis.
Under these conditions, it is not trivial to recognise splits and, most of all, merges, due to the amount of noise in the SAR image at time $t_2$. 
This case study highlights that an approach for change detection from an optical and a SAR image based on cluster splits and merges is limited by the clustering results.
These latter are affected, in turn, by the characteristics of the input data (noise ratio, contrast, etc.), by the adopted clustering algorithm, and by the selection of its hyperparameters.


\section{Conclusions and future works}
\label{sec:conclusions}

In this paper, we studied the challenging problem of change detection in multitemporal and heterogeneous images, by means of a cluster-based techniques. 
Our study focused on the case of image pairs, relative to the same area, captured by heterogeneous sensors at different times. 
We designed an unsupervised method, conceived to be application independent, which tackles a multitemporal image analysis problem, which is challenging due to the difficulty detecting changes from heterogeneous signals with no prior information.
We evaluated to which extent a completely unsupervised approach could be successful in addressing change detection from heterogeneous image sources.
Since there exist good models to capture the statistics of the kind of images considered (i.e.\ satellite optical and SAR images), the proposed method is based on distance measures that account for those properties.
The proposed idea is that changes on the ground can be related to clusters of the image at time $t_1$ splitting and/or merging into the clusters of the image at time $t_2$.
The possibility to model SAR intensity as log-normally distributed and optical data as Gaussian allowed us to apply a multivariate Gaussian model in the joint domain of the optical channels and of the log-transformed SAR data.
This allowed us to also apply the clustering algorithm on a stack of the images, to improve the chances of identify splits and merges.
Furthermore, a clustering algorithm has been developed that combines an extension of the fuzzy C-means algorithm with adaptive intra-cluster Mahalanobis metric and an ensemble approach aimed at minimizing dependence on the initialization.
Experimental results were obtained on real satellite heterogeneous images. 
The data set is relative to a flooded area and it contains pre-event and post-event images collected by optical and SAR spaceborne sensors. 
Our experiments demonstrated the relationship between the cluster splits and merges with the changed and unchanged areas.
This confirmed the potential of the clustering approach with respect to the problem of change detection from heterogeneous sources and suggested the effectiveness of the ensemble clustering approach.
However, the experiments also highlighted the limitations of this unsupervised approach.
In particular, the relationship between cluster splits/merges and changed/unchanged areas does not always hold.
This limitation can be addressed if prior, ancillary, or application-speciﬁc information is used to constrain the relationship between cluster splits/merges and changed/unchanged areas.
For example, possible improvements might result from:
\begin{itemize}
    \item providing some a priori information to the system, such as the most probable changing parts according to their position
    \item introducing hypothesis that the changing parts are the majority or the minority of the image, according to the particular application
    \item to indicate the particular class that represents the sought changed areas, e.g.\ water for floods, bare soil for forest ﬁres, etc.
\end{itemize}
The detection of cluster splits and merges carried out in this work is based on visual inspection and human interpretation, but it could be automated. 
A possible solution would be to overlap the mask of each cluster from the image at time $t_1$ to the stacked image, in order to identify the areas where splits occur. 
Accordingly, this procedure could be applied to the clusters of the image at time $t_2$ to recognise merges. 
Once splits and merges are identified, one could rely on prior information (if available) to improve the accuracy of change detection.
For example, if the location of the river in the image is provided, one could focus the search for flooded areas with in the clusters close to its position. 
Alternatively, one could leverage on the statistical characteristics of water in SAR images to identify the areas of interest.
On one hand, this approach would provide an automatic tool to improve clusters interpretation and to identify relevant splits and merge, associated with changes of interest.
On the other hand, the necessity of prior information confirms the extreme difficulty of performing automatic and unsupervised change detection in heterogeneous data.

A signiﬁcant improvement is also expected if polarimetric SAR images are used instead of SAR acquisitions with only one polarisation.
This is because they bring a lot more intrinsic information which would enhance the capability of clustering results to identify natural classes in feature spaces associated with SAR observations.
Obviously, this would force to consider different and more complicated models and distance measures.
The parallelisation of the proposed approach, which is favored by its window-based formulation and would beneﬁt of current cluster or GPU-based architectures, represents another possible and interesting future development.
Last, but not least, the research can be extended to the multitemporal case in which more than two images are considered, exploiting the proposed method for the analysis of long term trends such as deforestation, glacier dynamics, desertiﬁcation, land use change and urban development.
 
\bibliographystyle{unsrt}
\bibliography{Bibliography.bib}

\end{document}